\documentclass[sigconf]{acmart}
\AtBeginDocument{%
  }

\setcopyright{acmcopyright}
\copyrightyear{2023}
\acmYear{2023}
\acmDOI{XXXXXXX.XXXXXXX}

\acmConference[MM'23]{}{October 28, 2023 – November 3, 2023}{Ottawa, Canada}
\acmPrice{15.00}
\acmISBN{978-1-4503-XXXX-X/18/06}

\acmSubmissionID{1456}

\usepackage{graphicx}
\usepackage{amsmath}
\usepackage{booktabs}
\usepackage{xcolor}
\usepackage{bm}

\begin{document}

\renewcommand\footnotetextcopyrightpermission[1]{}
\settopmatter{printacmref=false} 
\title{Triple Correlations-Guided Label Supplementation for Unbiased Video Scene Graph Generation}

\author{Wenqing Wang\textsuperscript{1}\textsuperscript{\#} \quad Kaifeng Gao\textsuperscript{1}\textsuperscript{\#} \quad Yawei Luo \textsuperscript{1}\textsuperscript{*}   \quad Tao Jiang\textsuperscript{1} \quad Fei Gao\textsuperscript{2} \quad Jian Shao\textsuperscript{1} \\ \quad Jianwen Sun\textsuperscript{3} \quad Jun Xiao\textsuperscript{1}\\
\textsuperscript{1}Zhejiang University \quad
\textsuperscript{2}Zhejiang University of Technology \quad
\textsuperscript{3}Central China Normal University \\
\textsuperscript{\#}Co-first author \quad
\textsuperscript{*}Corresponding author\\
}

\renewcommand{\shortauthors}{Wang et al.}

\begin{abstract}
Video-based scene graph generation (VidSGG) is an approach that aims to represent video content in a dynamic graph by identifying visual entities and their relationships. Due to the inherently biased distribution and missing annotations in the training data, current VidSGG methods have been found to perform poorly on less-represented predicates. In this paper, we propose an explicit solution to address this under-explored issue by supplementing missing predicates that should be appear in the ground-truth annotations. Dubbed \textbf{Trico}, our method seeks to supplement the missing predicates by exploring three complementary spatio-temporal correlations. Guided by these correlations, the missing labels can be effectively supplemented thus achieving an unbiased predicate predictions. We validate the effectiveness of Trico on the most widely used VidSGG datasets, \emph{i.e.}, VidVRD and VidOR. Extensive experiments demonstrate the state-of-the-art performance achieved by Trico, particularly on those tail predicates. The code is available in the supplementary material.
\end{abstract}






\maketitle

\section{Introduction}
\label{sec:intro}
Video-based scene graph generation (VidSGG) targets at representing video content in the form of a dynamic graph constructed by \texttt{$\langle$subject, predicate, object$\rangle$} triplets. This structural representation makes VidSGG useful for downstream tasks like visual question answering~\cite{antol2015vqa,tapaswi2016movieqa,xiao2021next}, video captioning~\cite{xu2015show} and video retrieval~\cite{snoek2009concept,dong2021dual,wei2019neural}, \emph{etc}. Compared to its image-based counterpart, \emph{i.e.}, ImgSGG~\cite{zellers2018neural,misra2016seeing}, VidSGG is considered as a more challenging task since the pairwise relations between the visual entities are dynamic along the temporal dimension, making VidSGG a typical multi-label problem. These characteristics prevent those ImgSGG methods from being trivially applied to VidSGG despite the existence of a vast body of ImgSGG literature. In the current stage, VidSGG is a relatively under-explored problem and presents several unsolved issues.

Several early attempts have emerged to solve VidSGG by utilizing the spatio-temporal information of the video~\cite{qian2019video,teng2021target,cong2021spatial,liu2020beyond}. While these attempts have made some progress in improving overall performance by extracting short- and long-term information, they tend to ignore the inherent long-tail nature of the data, leading to severe bias in predicate predictions in the final results. It can be observed that most of the overall performance comes from a small number of head categories among all correctly detected predicates. Besides the inherent long-tail data distribution, another nature of VidSGG comes to the missing labels, which are inevitable during data annotation dues to the fleeting temporal interaction or inconspicuous spatial relation of the objects. Compared to the head predicates, the tail samples are more prone to be ignored by the annotators, which further deteriorates the predicate bias in a video.

\begin{figure}[t]
    \centering
    \includegraphics[width=\linewidth]{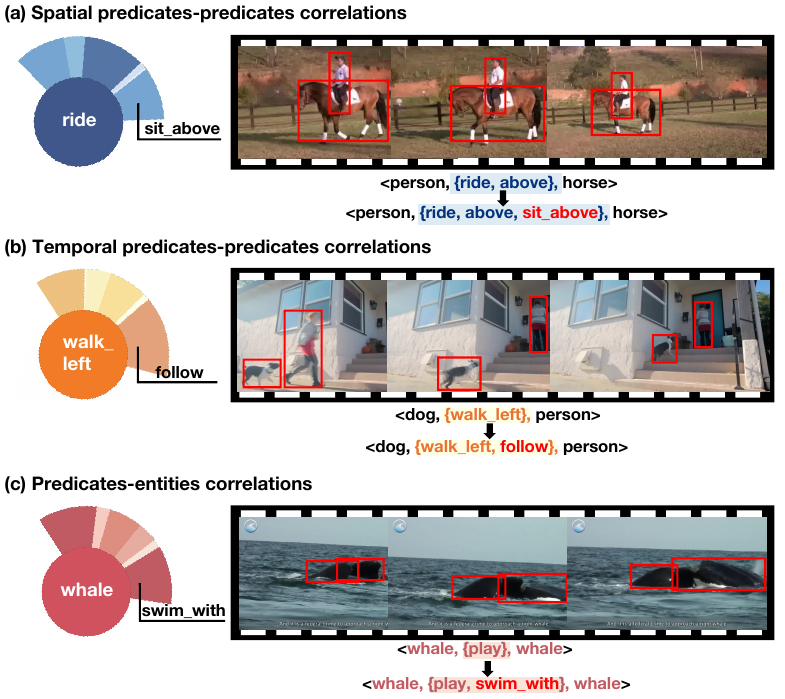} 
    \caption{
    \textbf{Examples of labels based on the Trico collaboration supplement.} 
    \textbf{Left}: The occurrence frequency of related predicates given a predicate or entity. \textbf{Right}: Examples of labels supplemented by the proposed Trico.}
    \label{fig:fig2}
    \vspace{-0.8cm}
 \end{figure}

More recently, a few methods~\cite{li2021interventional,xu2022meta} have attempted to address the biased predicate predictions in VidSGG. Li~\emph{et al.}~\cite{li2021interventional} proposed a causality-inspired interaction to weaken the false correlation between input data and the predicate labels. Xu \emph{et al.}~\cite{xu2022meta} considered temporal, spatial and object biases in a meta-learning paradigm. These \emph{implicit} approaches mitigate the long-tail problem to some extent, but the performance of tail classes is still not satisfied. 

In this paper, we propose a correlation-guided label supplementation method, dubbed \textbf{Trico}, to obtain an unbiased VidSGG model in an \emph{explicit} way. Concretely, we seek to supplement the missing predicates that are supposed to appear in the ground-truth annotations, by exploring the pairwise predicates-predicates as well as predicates-entities\footnote{Here, entities indicate objects or subjects.} correlations. These supplemented labels, together with the original ground truth, are then employed to train the model directly. Specially, we consider three kinds of correlations inherently existed in the original data to achieve label supplementation, including the spatial predicates-predicates correlations, temporal predicates-predicates correlations, and predicates-entities correlations. To demonstrate the necessity of exploring these triple correlations, we provide a motivating example in Figure~\ref{fig:fig2}.

The spatial predicates-predicates correlations hold dues to the spatial coherency, as shown in Figure~\ref{fig:fig2} (a), if \texttt{$\langle$person, ride, horse$\rangle$} occurs in a video, it is very likely that person is also \texttt{sit\_above} the horse. Similarly, temporal predicates-predicates correlations exist, as movements are coherent over time. For instance, if \texttt{$\langle$dog, walk\_left, person$\rangle$} happens in a video, it is highly probable that the dog would \texttt{follow} the person in the next video clip, as shown in Figure~\ref{fig:fig2} (b). Finally, predicates and entities are intertwined, as the choice of predicate depends on the specific entities involved. For example, if the subject is \texttt{whale} and the object is also \texttt{whale}, the predicate in this case probably goes to \texttt{swim\_with} beyond the commonplace predicate \texttt{play}, as shown in Figure~\ref{fig:fig2} (c). These three kinds of correlations provide strong cues for supplementing appropriate labels.

In practice, we utilize the triple correlations above in the form of correlation matrices, which are established by traversing the training dataset and calculating the conditional probability of the predicate given the concurrent predicates, previous predicates, or certain entities. The candidate labels are then generated elaborately by selecting predicates that are higher than the average predictive ability of the correlation-guided annotator model on each predicate. To enable the correlations and VidSGG model to be debiased mutually, the correlations are dynamically updated according to the prediction results of the target unbiased model trained on supplemented labels. 
In addition, we present a \textbf{logits smoothing} strategy to further improve the prediction ability of tail categories.

In summary, this paper makes the following contributions:
\begin{itemize}
\item We propose \textbf{Trico}, the first method to address VidSGG from an \emph{explicit} perspective of missing label supplementation.
\item We explore triple complementary correlations to guide the label supplementation. By capitalizing on the spatio-temporal cues offered by these correlations, the missing labels can be effectively supplemented to achieve an unbiased graph generation.
\item We verify the effectiveness of Trico on the most widely used VidSGG datasets, \emph{i.e.}, VidVRD~\cite{shang2017video} and VidOR~\cite{shang2019annotating}. Extensive results demonstrate the state-of-the-art performance achieved by Trico, especially on those tail predicates.
\end{itemize}

\section{Related Works}
\label{sec:relatedwork}

\subsection{Image-based Scene Graph Generation}
Image-based Scene Graph Generation (ImgSGG) is a task that involves describing objects and their relationships in an image. It derives various applications, such as image retrieval, image generation, image/video motion capture, and special relationship detection. Many methods have been proposed to tackle this task, and they can be divided into two main groups:

\textbf{(1) Two-stage methods:}~\cite{misra2016seeing,zellers2018neural,tang2019learning,chen2019knowledge}: In this method, object detection and relationship prediction are conducted within two stages. For example, Tang et al.~\cite{tang2019learning} first constructed a dynamic tree structure VCTree to place objects in the image in a visual environment to assist in visual reasoning tasks. Chen et al.~\cite{chen2019knowledge} integrated statistical correlation into deep neural networks and developed knowledge-embedded routing networks to promote scene graph generation.

\textbf{(2) One-stage methods}~\cite{liu2021fully,newell2017pixels,liao2020ppdm}: These methods deal with the object and relationship simultaneously in one go. However, due to the long-tail problem, the performances of traditional methods are far from satisfactory. Zeller~\emph{et al.}~\cite{zellers2018neural} first pointed out the imbalance of predicates in ImgSGG datasets. Chen~\emph{et al.}~\cite{chen2019knowledge} and Tang~\emph{et al.}~\cite{tang2019learning} also noticed this problem and proposed a new metric to measure the average performance, mean recall@K. Along this vein, various methods have been proposed to solve biased relationship prediction, including depolarization strategies such as resampling~\cite{li2021bipartite}, reweighting~\cite{yan2020pcpl}, and unbiased representation from bias~\cite{tang2020unbiased}, among others.


\subsection{Video-based Scene Graph Generation}
Generally, video-based and image-based SGG share a similar goal of detecting the visual objects and relationships in given data. Nevertheless, employing existing approaches designed for image-based SGG straightly to parse a video is non-trivial due to the multi-label and dynamic nature of VidSGG. Existing VidSGG work can be roughly divided into two groups according to the format of dataset annotation:

\textbf{(1) Frame-based VidSGG:} Ji~\emph{et al.}~\cite{ji2020action} proposed the first large-scale frame-level data set named \textbf{Action Genome (AG)}. Following this seminal work, researchers have proposed several methods that aim to capture complex spatio-temporal contextual information in videos for efficient VidSGG~\cite{cong2021spatial,teng2021target}. These methods include a new network structure called space-time converter ``(STTran)''~\cite{cong2021spatial}, and a frame-level VidSGG method termed ``TRACE''~\cite{teng2021target}, and a weakly-supervised VidSGG task with only single frame weak supervision ``SF-VidSGG’'~\cite{chen2023video}, among others. Both of these methods are designed for better capturing the spatio-temporal context information for relationship recognition.


\textbf{(2) Tracklet-based VidSGG:} Shang~\emph{et al.} proposed a commonly used three-stage segment-based detection framework in ~\cite{shang2017video}. However, this framework is not optimal for detecting relationships in long videos. To address this issue, researchers have proposed various methods to capture complex spatio-temporal contextual information in videos for efficient VidSGG~\cite{feng2021exploiting,liu2020beyond,woo2021and}. These methods include detection trajectory recognition paradigm~\cite{feng2021exploiting}, sliding window scheme ``VRD-STGC''~\cite{liu2020beyond}, and time span-suggested network ``TSPN''~\cite{woo2021and}, among others.
A more recent work by Shang~\emph{et al.}~\cite{shang2021video} introduced a method called VidVRD-II'' that incorporates iterative relational reasoning and joint relation classification. 
In our study, we use VidVRD-II''~\cite{shang2021video} and ``VRD-STGC''~\cite{liu2020beyond} as our base models and apply our Trico method to them. And Trico is evaluated on two video-level datasets, \textbf{VidVRD} and \textbf{VidOR} datasets, which aim to detect visual relationship instances in videos in the form of relational triplets \texttt{$\langle$subject, predicate, object$\rangle$} and object tracklets~\cite{shang2017video,shang2019annotating}. 

Based on our understanding, the work that is most similar to Trico is described in a paper by Xu et al. ~\cite{xu2022meta}. This method uses a meta-learning mechanism to address bias in both spatial and temporal aspects of video object segmentation. However, the method requires a complex process of splitting the dataset into support and query sets to simulate the distribution gap, which can be difficult to implement in practice. Trico, on the other hand, is different from Xu et al.'s method because it uses intrinsic correlations in the data to supplement missing labels and address bias in an explicit manner.


\section{Preliminaries}
\label{sec:method}

\subsection{Problem Formulation}

A video scene graph can be represented as $\mathcal{G} = (\mathcal{N},\mathcal{E})$ with an entity category set $\mathcal{C}_e$ and predicate category set $\mathcal{C}_p$. $\mathcal{N}$ and $\mathcal{E}$ are sets of nodes and edges, respectively. Each node in $\mathcal{N}$ has an entity category $c^e_i \in \mathcal{C}_e$ and a bounding box sequence (tracklet). Each edge in $\mathcal{E}$ has a linkage from the $i$-th node (subject) to the $j$-th node (object) and a collection of multiple predicate categories $\mathcal{P}_{ij}=\{c_k^p \in \mathcal{C}_p | k=1,\ldots,n_{ij}\}$, where $n_{ij}$ is the total number of predicates between the $i$-th and $j$-th node. Based on the above definition, our goal is to supplement labels $\mathcal{P}_{ij}$ for each $ij$-th subject-object pair by exploring predicate correlations.




\begin{figure*}[t!]
    \centering
    \includegraphics[width=\textwidth]{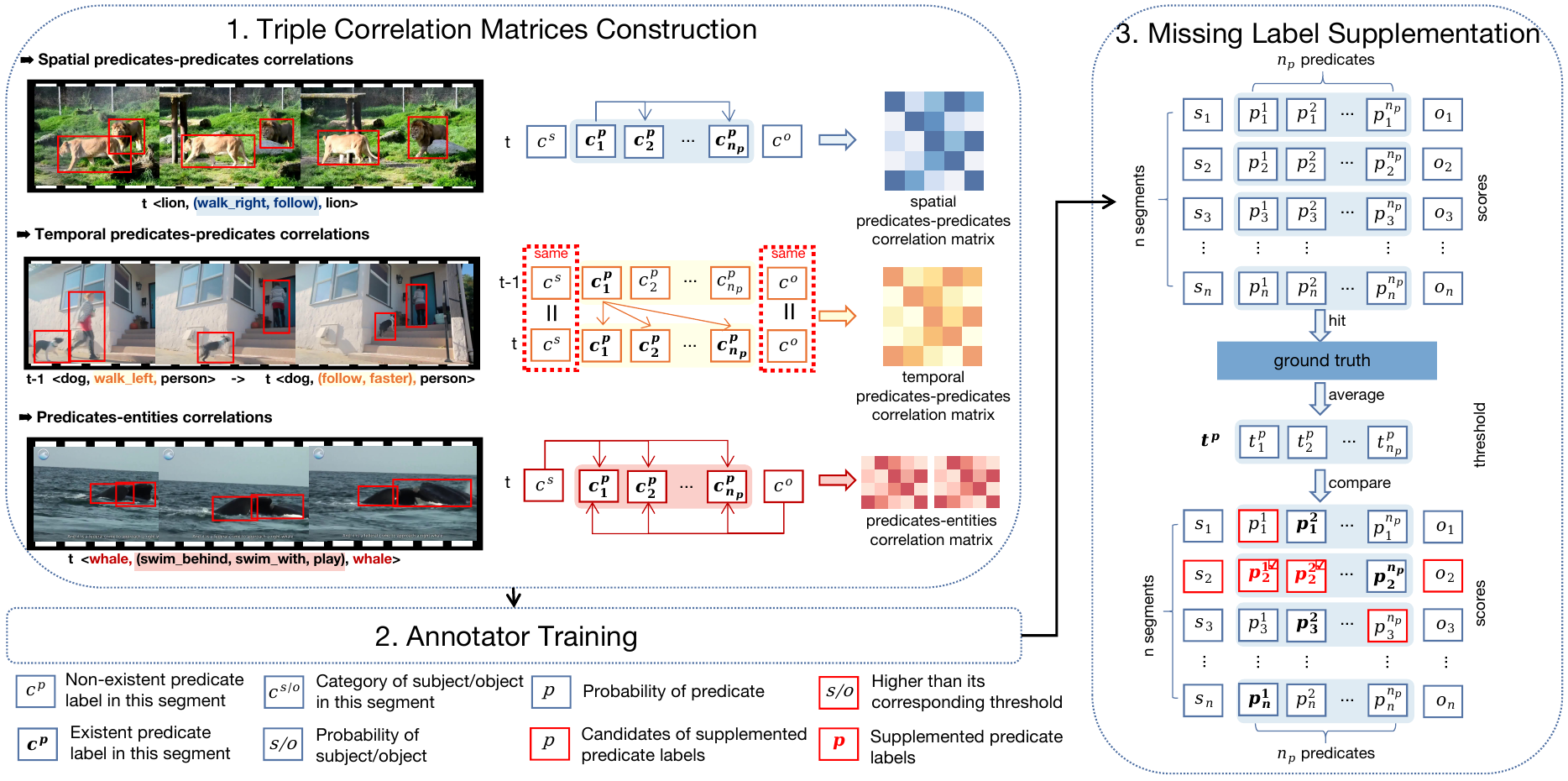} 
    \caption{\textbf{The overall framework of Triple Correlations-guided Missing Label Supplementation,} which includes Triple Correlation Matrices Construction, Annotator Training and Missing Label Supplementation.}
    \label{fig:main}
\vspace{-0.2cm}

 \end{figure*}

\subsection{Baseline Method}\label{basemethod}
A straightforward baseline method to characterize the predicate correlation is using the co-occurrence of predicate pairs. Specifically, we construct the correlation matrix $\bm{A}$ based on the joint probability of predicates $c^p_i$ and $c^p_j$ occurring in same video segment $\mathcal{V}$: 
\begin{equation}
\bm{A}(i,j) = P(c_i^p \in \mathcal{V}, c_j^p \in \mathcal{V}), ~\text{and}~  \bm{A} \in \mathbb{R}^{n_{p}\times{n_{p}}},
\end{equation}
where $n_{p}=|\mathcal{C}_p|$ is the number of predicate categories. Then we pre-train an auxiliary annotation model called ``Annotator'' by modifying the prediction based on the correlations in $\bm{A}$. \emph{I.e.}, predicate prediction $p_j$ is manually added a bias $\sum_{i=1}^{n_p}p_i\bm{A}(i,j)$ at the training time. Finally, we use the annotator to supplement predicate labels during inference on the training set.

The baseline method described above suffers from the long-tail problem of the dataset, which results in biased labels towards the head classes and numerous wrong labels being annotated. This is counterproductive to our original objective. Empirical evidence suggests that the baseline correlation works well for the head but not for the tail classes, as discussed in Section~\ref{sec:ablation}.

To address this issue and provide more accurate and unbiased label supplementation, we introduce our \textbf{Trico} method. Trico takes into account temporal information and considers three types of spatio-temporal correlations between predicates-predicates and predicates-entities to supplement labels that should have been annotated but were omitted.



\section{Methodology}
In this section, we present the formal introduction of \textbf{Trico}, which incorporates three spatio-temporal correlations to enhance the label supplementation process. Fig.~\ref{fig:main} shows our overall pipeline. We first construct the correlation matrices by considering the conditional probability of predicates-predicates and predicates-entities co-occurrence (Sec.~\ref{sec:construct_corr}). Subsequently, we pre-train an auxiliary annotation model called ``Annotator'' based on the constructed correlation matrices (Sec.~\ref{sec:reliable model}), and supplement predicate labels on the training set (Sec.~\ref{sec:supplement}). Finally, we train the target (unbiased) model to achieve the unbiased video relation detection (Sec. \ref{sec:FinalTrain}).

\subsection{Triple Correlation Matrices Construction}\label{sec:construct_corr}
We consider three types of correlations that are complementary and include both spatio-temporal information.

\noindent\textbf{Spatial Predicates-Predicates Correlations.} 
Many predicates occur conditionally given another predicate occurring in a certain spatial layout. To take advantage of such spatial cues, we construct the spatial predicates-predicates correlations matrix $\boldsymbol{A}_{S} \in \mathbb{R}^{n_{p}\times{n_{p}}}$ by considering the conditional probability of predicate $c_j^p$ given the predicate $c_i^p$:
\begin{equation}
    \bm{A}_S(i,j) = P(c_j^p \in \mathcal{V} | c_i^p \in \mathcal{V}).
\end{equation}

\noindent\textbf{Temporal Predicates-Predicates Correlations.} Due to the temporal continuity of videos, previous video segments can provide strong cues for supplementing missing predicates in the current segment. Accordingly, we construct the temporal matrix $\bm{A}_{T} \in \mathbb{R}^{n_{p}\times{n_{p}}}$ by considering the conditional probability of $c_j^p$ in the current segment given the occurrence of $c_i^p$ in the previous segment, \emph{i.e.}, 
\begin{equation}
    \bm{A}_T(i,j) = P(c^p_j \in \mathcal{V}_{cur} | c^p_i \in \mathcal{V}_{pre} ),
\end{equation}
where $\mathcal{V}_{cur}$ is the current video segment, and $\mathcal{V}_{pre}$ is the previous video segment. To capture more precise correlations, we only count the occurrence of predicates within the same subject-object pair across the two segments.

\noindent\textbf{Predicates-Entities Correlations.} The intertwined correlations between entities and predicates also offers insight for the possible missing labels. Based on this thought, we construct the predicates-entities correlations matrix $\boldsymbol{A}_{E} \in \mathbb{R}^{n_{e}\times{n_{p}}}$ (where $n_{e}=|\mathcal{C}_e|$). Similarly, we calculate the conditional probability of predicate $c_j^p$ given the occurrence of entity $c_i^e$ as subject or object:
\begin{equation}
    \bm{A}_E(r, i,j) = P(c_j^p \in \mathcal{V} | c_i^e \in \mathcal{V}, c_i^e~ \text{as} ~r),
\end{equation}
where $r \in \{\text{s,o}\}$ represents the semantic roles ( \emph{i.e.}, subject or object) of the entity, each channel of $\bm{A}_E$ represents the correlation of subject and object.

\subsection{Annotator Training}\label{sec:reliable model}
In order to train a less biased model (called annotator) for missing label supplementation, we use the constructed correlation matrices to adjust the predicate predictions. Specifically, for each subject-object tracklet pair, we consider the model prediction (\emph{i.e.}, the predicate classification probability $p_j$) and the correlation prior (denoted as $q_j$) as multivariate Bernoulli distribution. The desired annotating output is at least one of them (\emph{i.e.}, either $p_j$ or $q_j$) is correct. Thus, we adjust the predictions of the 
initial model as 
\begin{equation}
    \hat{p}_{j} = 1-(1-p_{j})(1-q_j), ~ j=1,\ldots, n_p \label{eq:general},
\end{equation}
where $(1-p_j)$ and $(1-q_j)$ represent the incorrect probability of the initial model and the correlation prior, respectively. We then train the annotator based on the updated prediction $\hat{p}_j$. Following VidVRD-II~\cite{shang2021video}, we use binary cross-entropy (BCE) loss and train the predicate classification branch together with entity classification.

To consider different correlations, we model the prior probability $q_j$ \emph{w.r.t} three types of correlations. For simplicity, we define the function $F_*(\cdot)$ as the \emph{incorrect} prior probability (\emph{i.e.}, $1-q_j$) for each type of correlation:

\noindent\textbf{Function of Correlation $\bm{A}_S$.} We consider correlation effect of predicate $j$ conditioned on all other predicates. The probability $q_{j,i}$ is contributed by the predicate $i$ based on the correlation prior, \emph{i.e.}, $q_{i,j} = p_{i} \boldsymbol{A}_{S}{(i,j)}$. 
By considering all of the prior predictions $\{q_{i,j}\}$ excluding $i=j$, the incorrect prior probability can be calculated as 
\begin{equation} 
F_S(j)={\textstyle \prod}_{i=1,i \neq j}^{n_p}(1-p_{i}\boldsymbol{{A}}_S{(i,j)}).\label{eq:spatio}
\end{equation}

\noindent\textbf{Function of Correlation $\bm{A}_T$.} Similarly, we can calculate the incorrect prior from $\bm{A}_T$ by considering the prediction of predicate $i$ in the previous video segment:
\begin{equation} 
F_T(j) = {\textstyle \prod}_{i=1}^{n_p}(1-p_{i,pre}\bm{A}_T{(i,j)}), \label{eq:temporal}
\end{equation}
where $p_{i,pre} $ denotes the probability of $i$-th predicate in the previous segment.

\noindent\textbf{Function of Correlation $\bm{A}_E$.} Finally we compute the correlation prior from $\bm{A}_E$ and the entity predictions, considering both cases where the entity serves as subject or object from the current predicate $j$. Concretely, the incorrect prior probability function is
\begin{equation} 
F_E(j) =  {\textstyle \prod}_{r \in \{\text{s,o}\}}{\textstyle \prod}_{i=1}^{n_e}(1-p^r_i \bm{A}_E(r,i,j)),
\end{equation}  %
where $p^r_i$ denotes the probability of $i$-th entity category and serves as different semantic roles (\emph{i.e.}, subject or object).


\subsection{Missing Label Supplementation}\label{sec:supplement}

After obtaining the annotator model, we use it to supplement predicate labels for each training sample through an inference process.
To ensure high-quality labels, we select candidate samples that 1) match the ground-truth subject-object pair, and 2) have an entity classification score higher than the average hitting score.

For each candidate sample (\emph{i.e.}, subject-object pair), we collect the predicate predictions based on each type of correlation (cf. Eq. (\ref{eq:general})), and calculate the predicate confidence thresholds $\bm{t}_S,\bm{t}_T,\bm{t}_E \in [0,1]^{|\mathcal{C}_p|}$ respectively by averaging $p_j^p$ from all samples for each category. Consequently, for each sample (\emph{i.e.}, subject-object pair $ij$), the supplemented predicate labels are
\begin{equation}
   \mathcal{P}_{ij}^S \cup \mathcal{P}_{ij}^T \cup \mathcal{P}_{ij}^E,  ~\text{where}~ \mathcal{P}_{ij}^*= \{c^p_k|p_k > \bm{t}_*(k)\}.\label{eq:labelSupp}
\end{equation}

\subsection{Target (Unbiased) Model Training}\label{sec:FinalTrain}
In this step, we combine the additional labels (\emph{i.e.}, Eq. (\ref{eq:labelSupp})) obtained from the annotator with the original ground-truth labels to train the final target model. The target model has the same network structure as the annotator, and we use VRD-STGC~\cite{liu2020beyond} and VidVRD-II~\cite{shang2021video} as the relation detection pipeline. The objective function for training the target model is also chosen as a binary cross-entropy (BCE) loss, which is similar to the one used for training the annotator.

\noindent\textbf{Dynamic Correlation Updating.} During the training process of the target model, we update the correlation matrices based on the latest prediction results from the model. This allows the correlation matrices and the VidSGG model to be debiased in an iterative manner. We use a moving average strategy to fine-tune the matrices during training. Specifically, we calculate the updated correlation matrices as a weighted average of the current matrix and the previous matrix, where the weight is determined by a decay factor. This allows the matrices to adapt to the changing patterns of the data and converge to a more accurate representation. We then use the updated matrices to re-conduct the annotator and supplement missing labels for the target model training. The moving average strategy is as follows:
\begin{equation} 
\boldsymbol{A}^{(t)}_* = \eta\boldsymbol{A}^{(t)}_* + (1-\eta) \boldsymbol{A}^{(t-1)}_*,
\end{equation}
where $\bm{A}_* \in \{\boldsymbol{A}_S, \boldsymbol{A}_T, \boldsymbol{A}_E\}$, $t \geq 1$ indexes the training epoch, and $\eta$ is the hyper-parameter of moving average. Note that $\bm{A}^{(0)}_*$ is the initial correlation matrix constructed by making statistics on the training set, and $\bm{A}^{(t)}_*$ ($t\geq 1$) is constructed by the prediction results of the target model.

\noindent\textbf{Logits Smoothing.}
To balance the learning ability of the model between head and tail predicates, we propose a strategy called logits smoothing. This involves reducing the prediction logits of tail predicates during training, which are less frequently represented in the training data, in order to shift the training focus towards them by strengthening the loss value of the tail labels. We use $\mathcal{M}$ to represent the smoothing parameters, which are initialized based on the distribution of all predicate categories, and then scaled using hyper-parameters $\alpha$ and $\beta$:
\begin{equation}
\mathcal{M} \Leftarrow {\beta \mathcal{M} ^\alpha}/{\max(\mathcal{M} ^\alpha)}.\label{eq:smoothing}
\end{equation}
During training, we smooth the corresponding prediction logits of predicates $p_j$ by $p_j - \mathcal{M}(j)$. This helps to balance the model's learning between head and tail predicates and transfer the training focus to the less represented tail predicates. Our proposed logits smoothing strategy is simple yet effective and has been found to be helpful for exploring the tail information.

\section{Experiments}
\label{sec:experiments}


\subsection{Datasets}
We conducted experiments on two video relation benchmarks: ImageNet-VidVRD~\cite{shang2017video} and VidOR~\cite{shang2019annotating}:

\noindent\textbf{ImageNet-VidVRD}~\cite{shang2017video} is the first dataset used for benchmarking VidSGG and has been widely used in previous research. It consists of $1,000$ videos (with $800$ videos for training and $200$ videos for evaluation) from ILSVRC2016-VID and is manually annotated with video relation instances. The dataset covers $35$ categories of subjects/objects and $132$ categories of predicates in total. 

\noindent\textbf{VidOR}~\cite{shang2019annotating} is a large-scale dataset of user-generated videos collected from social media. It covers $80$ categories of subjects/objects and $50$ categories of predicates in total. VidOR consists of a training set with $7,000$ videos, a validation set with $835$ videos, and a testing set with $2,165$ videos. We evaluated our approach using the validation set.

\subsection{Evaluation Tasks and Metrics}\label{sec:mr}

\noindent{\textbf{Tasks:}} We follow some standard tasks of ImgSGG~\cite{xu2017scene} for evaluation: (1) Predicate Classification \textbf{(PredCls)}, where the goal is to predict the predicate classes given the ground-truth labels and bounding boxes of objects, and (2) Scene Graph Detection \textbf{(SGDet)}, where the goal is to detect the labels and bounding boxes of objects, and predict the predicate classes. 

\noindent{\textbf{Metrics:}} We use the official evaluation metrics~\cite{shang2017video,shang2021video} of the VRU Challenge, including Relation Detection (RelDet) and Relation Tagging (RelTag). We report the average precision \textbf{(mAP)}, Recall@K (\textbf{R@K},K=$50,100$) for RelDet and Precision@K (\textbf{P@K},K=$5,10$) for RelTag. In addition, following the approach of~\cite{chen2019knowledge,tang2019learning,tang2020unbiased} and~\cite{li2022label} we introduce the mean Recall@K \textbf{(mR@K)} and \textbf{Mean} as key metrics for evaluating VidSGG. 

\noindent{\textbf{mR@K:}} mR@K is calculated by first obtaining recall scores from the top-K triplet predictions in every video segment and then averaging them \emph{w.r.t} each predicate category. As discussed in previous work such as~\cite{chen2019knowledge,tang2019learning,tang2020unbiased}, mR@K is considered a more canonical metric in scenarios with imbalanced data. Therefore, the design of Trico mainly focuses on optimizing mR@K.

\noindent{\textbf{Mean:}} As discussed in ~\cite{li2022label}, Mean is calculated as the average of mR@K and R@K. Since R@K tends to favor head predicates and mR@K tends to favor tail predicates, Mean metric provides a more balanced performance evaluation across all (head and tail) predicates. \textbf{Accordingly, we use ``Mean'' as our primary evaluation metric in our experiments.}

\subsection{Implementation Details}

\noindent{\textbf{Relation Detection Details.}}
We provide details on how we perform relation detection for the two datasets used in our experiments: VRD-STGC~\cite{liu2020beyond} and VidVRD-II~\cite{shang2021video}.
\textbf{(1) VRD-STGC}~\cite{liu2020beyond}: We use a Faster-RCNN model to detect objects in each video frame, and then track the detected objects across the entire video using a Multiple Object Tracking (MOT) algorithm to obtain tracklets. For each object pair, we extract RoI-aligned detection and I3D features with relative motion features. We only consider pairs whose bounding boxes overlap with the ground truth by more than 0.5 in volume IoU (vIoU). We then construct a spatial graph and a temporal graph to filter out incompatible proposals.
\textbf{(2) VidVRD-II}~\cite{shang2021video}: We use a shot segmentation technique to split the video into segments of 30 frames with 15 frames overlapping. FasterRCNN~\cite{ren2015faster} is used to detect bounding boxes in each frame, and Seq-NMS~\cite{han2016seq} is applied to generate tracklets (i.e., sequences of bounding boxes). The pipeline then employs RoI Aligned visual features of tracklet regions and the relative position feature of subject-object pairs to classify relations. Finally, the detected relation instances are associated across segments using a simple greedy relation association algorithm proposed in~\cite{shang2017video}.

\begin{table*}[t!]
\centering
\caption{Performance (\%) of Trico and other baselines on VidVRD~\cite{shang2017video} dataset in PredCls.  \textbf{LoS}: Logits smoothing strategy, \emph{i.e.}, Eq.~(\ref{eq:smoothing}). \textbf{Mean}: The average of mR@50/100 and R@50/100.
}
    \label{tabel:vidvrd-predcls}
\resizebox{0.75\linewidth}{!}{ 
\begin{tabular}{c|cc|cc|c|c|cc}
\hline & \multicolumn{6}{c|}{Relation Detection}            & \multicolumn{2}{c}{Relation Tagging} \\ Method & mR@50 & \multicolumn{1}{c|}{mR@100} & R@50  & R@100 & Mean & mAP      & P@5        & P@10       \\ \hline
VidVRD-II~\cite{shang2021video}                 & 37.09 & 45.45   & 44.43 & 59.28  & 46.56 & 47.32  & 47.30 & 36.50 \\
\textbf{+Trico (ours)}    & 36.57 & 48.10   & \textbf{44.63} & \textbf{59.65}  & 47.24 & \textbf{48.47}  & 48.20 & 36.40\\
\textbf{+Trico+LoS (ours)} & \textbf{37.21} & \textbf{49.15}   & 43.35 & 59.38   & \textbf{47.27} & 48.17  & \textbf{48.60} & \textbf{37.45}\\ \hline
\end{tabular}
}
\vspace{-5pt}
\end{table*}

\begin{table*}[t!]

\centering
\caption{Performance (\%) of Trico and other baselines on VidOR~\cite{shang2019annotating} dataset in PredCls. \textbf{S, E, T}: using correlation of $\bm{A}_S,\bm{A}_E,\bm{A}_T$ respectively. \textbf{Mean}: The average of mR@50/100 and R@50/100.
}
    \label{tabel:vidor-predcls}
\resizebox{0.75\linewidth}{!}{ 
    \begin{tabular}{c|cc|cc|c|c|cc}
    \hline & \multicolumn{6}{c|}{Relation Detection}            & \multicolumn{2}{c}{Relation Tagging} \\ Method & mR@50 & \multicolumn{1}{c|}{mR@100} & R@50  & R@100 & Mean & mAP      & P@5        & P@10       \\ \hline
VidVRD-II~\cite{shang2021video}                 & 23.37 & 29.75   & \textbf{52.06} & \textbf{68.94} & 43.53 & \textbf{62.11}  & 42.50 & \textbf{32.74} \\
\textbf{+S (ours)}    & 23.24 & 30.13   & 51.73 & 68.39 & 43.37  & 61.49  & \textbf{42.96} & 32.37 \\
\textbf{+E (ours)}    & 23.30 & 30.19   & 51.94 & 68.70 & 43.53  & 61.69  & 42.48 & 32.14 \\
\textbf{+T (ours)}    & 24.43 & 31.37   & 50.36 & 66.88 & 43.26  & 57.97  & 39.45 & 30.58 \\
\textbf{+Trico (ours)}    & 25.00 & \textbf{32.07}   & 50.38 & 67.43  & \textbf{43.72} & 57.84  & 40.34 & 30.93 \\
\textbf{+Trico+LoS (ours)}    & \textbf{25.02} & 30.83   & 50.85 & 67.46  & 43.54 & 59.65  & 41.03 & 31.42 \\ \hline
\end{tabular}
}
\vspace{-5pt}
\end{table*}

\begin{table}[t!]

\centering
\caption{Performance (\%) on VidVRD~\cite{shang2017video} dataset in SGDet. \textbf{Mean}: The average of mR@50/100 and R@50/100.} 
\label{tabel:vidvrd-sgdet}
\tabcolsep = 0.06cm
\resizebox{\linewidth}{!}{
    \begin{tabular}{c|cccc|c|c|cc}
    \hline
     & \multicolumn{6}{c|}{Relation Detection}            & \multicolumn{2}{c}{Relation Tagging} \\ Method
                            & mR@50 & \multicolumn{1}{c|}{mR@100} & R@50  & R@100 & Mean & mAP      & P@5        & P@10       \\ \hline
    VidVRD~\cite{shang2017video}                  & -     & \multicolumn{1}{c|}{-}      & 5.54  & 6.37  & - & 8.58  & 28.90      & 20.80      \\
    GSTEG~\cite{tsai2019video}                   & -     & \multicolumn{1}{c|}{-}      & 7.05  & 8.67  & - & 9.52  & 39.50      & 28.23      \\
    3DRN~\cite{20213}                    & -     & \multicolumn{1}{c|}{-}      & 5.53  & 6.39  & - & 14.68 & 41.80      & 29.15      \\
    VRD-GCN~\cite{qian2019video}        & -     & \multicolumn{1}{c|}{-}      & 8.07  & 9.33  & - & 16.26 & 41.00      & 28.50      \\
    MHA~\cite{su2020video}              & -     & \multicolumn{1}{c|}{-}      & 9.53  & 10.38 & - & 19.03 & 41.40      & 29.45      \\
    TRACE~\cite{teng2021target}                   & -  & \multicolumn{1}{c|}{-}   & 9.08  & 11.15 & - & 17.57 & 45.30      & 33.50      \\
    TSPN~\cite{woo2021and}                    & -     & \multicolumn{1}{c|}{-}      & 11.56 & 14.13 & - & 18.90 & 43.80      & 33.73      \\
    Social Fabric~\cite{Chen2021Social}          & -     & \multicolumn{1}{c|}{-}      & 13.73 & 16.88 & - & 20.08 & 49.20      & 38.45      \\
    IVRD~\cite{li2021interventional}                    & -     & \multicolumn{1}{c|}{-}      & 12.40 & 14.46 & - & 22.97 & 49.87      & 35.75      \\ \hline
    VRD-STGC~\cite{liu2020beyond}                & \textbf{8.73}  & \multicolumn{1}{c|}{10.21}  & 11.21 & 13.69 & 10.96 & \textbf{18.38} & \textbf{43.10}      & \textbf{32.24}      \\
    \textbf{+Trico (ours) }              & 8.69 & \multicolumn{1}{c|}{\textbf{10.80}}  & \textbf{12.18} & \textbf{15.10} & \textbf{11.69} & 16.22  & 40.40      & 31.31      \\
    \hline
    VidVRD-II~\cite{shang2021video}               & 12.36 & \multicolumn{1}{c|}{13.33}  & \textbf{13.38} & \textbf{14.93} & 13.50 & \textbf{25.10} & 52.80      & \textbf{40.05}      \\ 
    \textbf{+Trico (ours) }              & \textbf{12.55} & \multicolumn{1}{c|}{13.61}  & 13.09 & 14.64 & 13.47 & 24.25  & \textbf{53.10}      & 39.70      \\
    \textbf{+Trico+LoS (ours)}               & 12.52 & \multicolumn{1}{c|}{\textbf{13.78}}  & 13.24 & 14.71 & \textbf{13.56} & 24.10  & 52.30      & 39.75      \\ \hline
    \end{tabular}
}
\vspace{-13pt}
\end{table}

\noindent{\textbf{Hyper-Parameters.}} 
\textbf{(1) VRD-STGC~\cite{liu2020beyond}}: To update the dynamic correlation on VidVRD, we update $\bm{A}_S, \bm{A}_T, \bm{A}_E$ every $10$ epochs with a learning rate of $\eta=$ 1e-4. We train our model using SGD for a total of $20$ epochs with a learning rate of 1e-1.
\textbf{(2) VidVRD-II~\cite{shang2021video}}: To update the dynamic correlation on VidVRD, we update $\bm{A}_S$ every $15$ epochs with a learning rate of $\eta=$ 1e-5, $\bm{A}_T$ every $15$ epochs with a learning rate of $\eta=$ 1e-4, and $\bm{A}_E$ every $5$ epochs with a learning rate of $\eta=$1e-4. For logits smoothing, we set $\alpha=-0.25$ and $\beta=40$ in all experiments. Other hyper-parameters are set consistently with VidVRD-II~\cite{shang2021video}. We train our model for a total of $50$ epochs with a learning rate of 1e-3 using Adam~\cite{kingma2014adam}.

\begin{table}[t!]
\centering
\caption{Performance (\%) on VidOR~\cite{shang2019annotating} dataset in SGDet. \textbf{Mean}: The average of mR@50/100 and R@50/100.} 
\label{tabel:vidor-sgdet}
\tabcolsep = 0.06cm
\resizebox{\linewidth}{!}{
	\begin{tabular}{c|cccc|c|c|cc}
    \hline
     & \multicolumn{6}{c|}{Relation Detection}                     & \multicolumn{2}{c}{Relation Tagging} \\ Method
                            & mR@50 & \multicolumn{1}{c|}{mR@100} & R@50  & R@100 & Mean & mAP      & P@5        & P@10       \\ \hline
    RELAbuilder~\cite{zheng2019relation}                  & -     & \multicolumn{1}{c|}{-}      & 1.58  & 1.85  & - & 1.47  & 35.27      & -      \\
    3DRN~\cite{20213}                  & -     & \multicolumn{1}{c|}{-}      & 2.58  & 2.75  & - & 2.47  & 42.33      & 29.89      \\
    MaGUS.Gamma~\cite{sun2019video}                 & -     & \multicolumn{1}{c|}{-}      & 6.89  & 8.83  & - & 6.56  & 40.73      & -      \\
    MHA~\cite{su2020video}                    & -     & \multicolumn{1}{c|}{-}      & 6.35  & 8.05  & - & 6.59  & 41.56      & 32.53      \\
    VRD-STGC~\cite{liu2020beyond}        & -     & \multicolumn{1}{c|}{-}      & 8.21  & 9.90  & - & 6.84  & 36.78      & -      \\
    IVRD~\cite{li2021interventional}                & -  & \multicolumn{1}{c|}{-}  & 7.36 & 9.41 & - & 7.42 & 42.70      & -      \\
    TSPN~\cite{woo2021and}              & -     & \multicolumn{1}{c|}{-}      & 9.33  & 10.71  & - & 7.61  & 42.22      & 34.94      \\
    Social Fabric~\cite{Chen2021Social}             & 6.74     & \multicolumn{1}{c|}{7.71}      & 9.99  & 11.94 & 9.03 & 11.21  & 55.16      & -      \\\hline
    VidVRD-II~\cite{shang2021video}               & 2.70 & \multicolumn{1}{c|}{2.83}  & 4.32 & 5.07 & 3.73 & 4.31  & \textbf{44.45}      & 33.59      \\ 
    \textbf{+S (ours) }              & 3.17 & \multicolumn{1}{c|}{3.32}  & 4.38 & 5.12 & 4.00 & \textbf{4.57}  & 44.18      & \textbf{33.63}      \\ 
    \textbf{+E (ours) }              & 2.87 & \multicolumn{1}{c|}{3.01}  & \textbf{4.41} & 5.13 & 3.86 & \textbf{4.57}  & 44.01      & 33.49      \\ 
    \textbf{+T (ours) }              & 3.24 & \multicolumn{1}{c|}{3.43}  & 4.19 & 4.97 & 3.96 & 4.09  & 41.78      & 32.30      \\ 
    \textbf{+Trico (ours) }              & 3.39 & \multicolumn{1}{c|}{3.60}  & 4.14 & 4.98 & 4.03 & 4.06 & 42.50      & 32.43      \\
    \textbf{+Trico+LoS (ours)}               & \textbf{3.44} & \multicolumn{1}{c|}{\textbf{3.64}}  & 4.36 & \textbf{5.22} & \textbf{4.17} & 4.24  & 43.32      & 33.13      \\ \hline
    \end{tabular}
}
\vspace{-13pt}
\end{table}

\begin{table}[t!]
\centering
\caption{Ablation (\%) on VidVRD~\cite{shang2017video} dataset in PredCls. \textbf{Base}: The baseline correlation (Sec. \ref{basemethod}). \textbf{+T+S+E}: Our method Trico.}
    \label{tabel:ablation}
    \tabcolsep = 0.06cm
\resizebox{\linewidth}{!}{
\begin{tabular}{c|c|ccc|c}
\hline
 &        & Head   & Body   & Tail   &     \\
 Method  & mR@100 & mR@100 & mR@100 & mR@100 & mAP \\ \hline
\multicolumn{1}{c|}{VidVRD-II~\cite{shang2021video}} & 45.35 & 78.08 & 49.55 & 30.36 & 47.49  \\
\multicolumn{1}{c|}{+base} & 45.84 & \textbf{78.83} & 54.82 & 28.24     & 46.15 \\\textbf{+T (ours)} & 47.66 & 78.33 & 51.52 & 33.65     & 46.43  \\\textbf{+T+S (ours)} & 48.04 & 74.74 & 53.60 & \textbf{34.70}     & 47.40 \\\textbf{+T+S+E (ours)} & \textbf{48.10} & 76.46 & \textbf{56.90} & 32.77    & \textbf{48.47}  \\ \hline
\end{tabular}
}
\vspace{-5pt}
\end{table}

\subsection{Comparison with SOTAs}

\noindent{\textbf{Performance Comparison on VidVRD.}} In Table~\ref{tabel:vidvrd-predcls}, Trico is compared with VidVRD-II~\cite{shang2021video} on the PredCls task of VidVRD. For fair comparison, VidVRD-II is re-implemented in our codebase. Trico achieves a significant improvement of 2.65\% in mR@100 on the PredCls task, and slight improvements under other metrics like mAP. Compared to existing methods, Trico only slightly decreases under R@K, but increases by 0.71\% in Mean, a metric that takes into account all predicates, indicating that Trico does not sacrifice much R@K to improve mR@K. Equipping Trico with the logits smoothing strategy results in improvements under all metrics on the PredCls task, showing its effectiveness in dealing with the long-tail problem in predicate distribution.

Additionally, Trico is compared with other SOTA methods on the SGDet task in Table~\ref{tabel:vidvrd-sgdet}. Trico is applied to both the VRD-STGC and VidVRD-II models, resulting in significant improvements. For the VRD-STGC model, Trico achieves a 0.59\% improvement in mR@100, 0.97\%/1.41\% improvement in R@50/100, and 0.73\% improvement in Mean. For the VidVRD-II model, Trico achieves a 0.19\%/0.28\% improvement in mR@50/100 and a 0.16\%/0.45\% improvement in mR@50/100 when equipped with the logits smoothing strategy.

\begin{figure*}[t!]
    \centering
    \includegraphics[width=\linewidth]{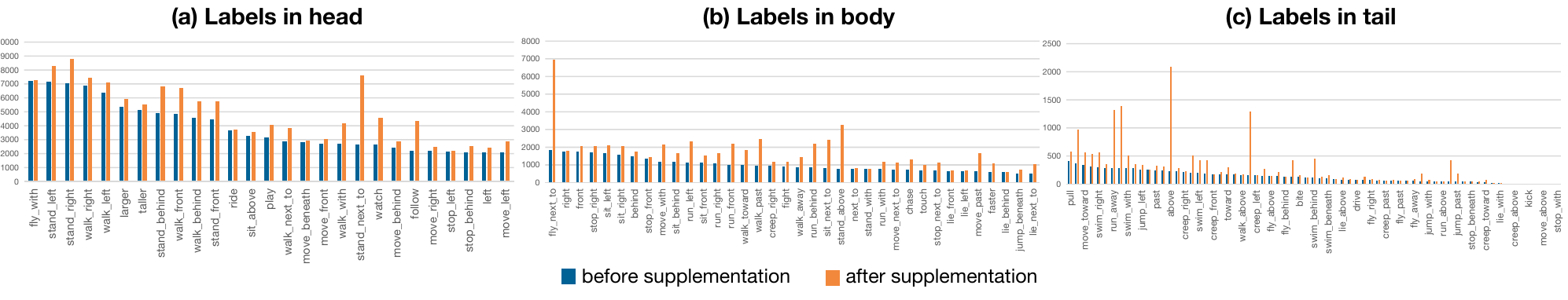} 
    \caption{\textbf{Statistical results of label supplementing on VidVRD}. \textcolor[rgb]{0.1914,0.3867,0.578}{Blue}: labels before supplementation.  \textcolor[rgb]{0.9375,0.492,0.1719}{Orange}: labels after supplementation.}
    \label{fig:Distribution}
\vspace{-0.2cm}

 \end{figure*}
 
\begin{figure*}[t!]
    \centering
    \includegraphics[width=\textwidth]{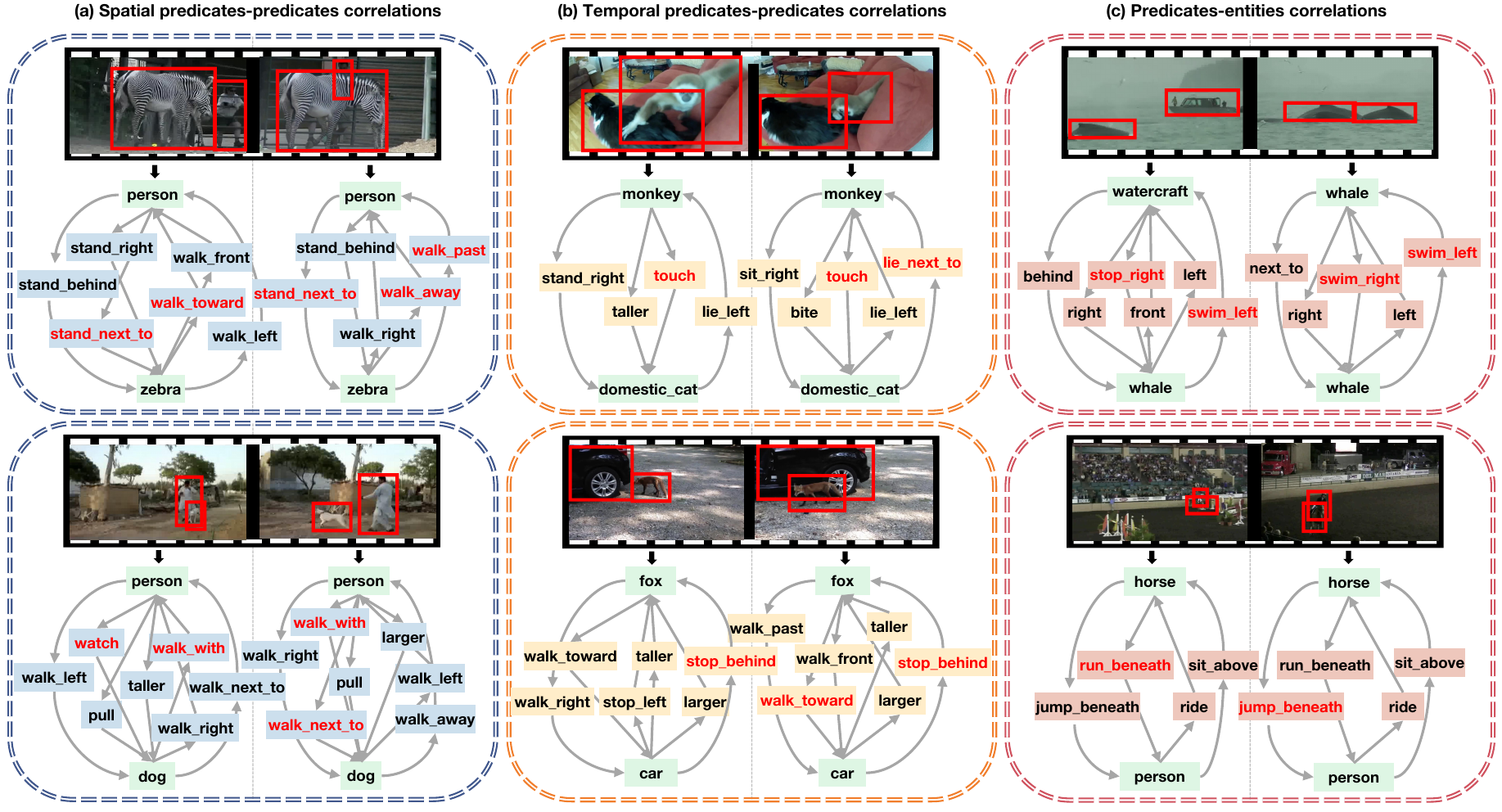} 
    \caption{\textbf{Visualization of the generated video scene graph on VidVRD dataset.} In the same box, we build a video scene graph of two adjacent segments in the same video. We have supplemented labels from three kinds of correlations: $\bm{A}_S$ in (a), $\bm{A}_T$ in (b), and $\bm{A}_E$ in (c). The \textcolor{red}{\textbf{red labels}} are supplemented labels and the \textbf{black labels} are the labels before supplementation.}
    \label{fig:Results}
    \vspace{-0.3cm}
\end{figure*}

\noindent{\textbf{Performance Comparison on VidOR.}} In Table~\ref{tabel:vidor-predcls}, we present a comparison between Trico and VidVRD-II~\cite{shang2021video} on the PredCls task for VidOR. Trico achieves a significant improvement in mR@100, increasing it from 29.75\% to 32.07\%, which demonstrates its effectiveness in enhancing the average predicate category capacity. Using $\bm{A}_S$, $\bm{A}_E$, or $\bm{A}_T$ alone results in comprehensive improvements, while using $\bm{A}_T$ alone harms the mAP performance (-4.27\%), leading to unsatisfactory mAP results. We speculate that using $\bm{A}_T$ alone would introduce too many noisy labels. However, after applying the logits smoothing strategy, this situation is alleviated.

Furthermore, in Table~\ref{tabel:vidor-sgdet}, we compare Trico with other SOTA methods on the SGDet task for VidOR. Trico achieves better performance in mR@K and R@K for $\bm{A}_S$ and $\bm{A}_E$ compared to the baseline model. However, using $\bm{A}_T$ alone results in a slight decrease in R@K.

\subsection{Ablation Study}\label{sec:ablation}
We conducted ablation studies to demonstrate the importance of the three correlations in Trico, as shown in Table~\ref{tabel:ablation}. To provide a more detailed comparison, we divided the predicate categories into three groups based on their occurrence frequency: Head (27), Body (36), and Tail (69) (cf. Figure \ref{fig:Distribution}).

From Table~\ref{tabel:ablation}, we can observe that the baseline method in the second row strengthens the ability of the head labels while making the ability of the tail labels even worse, which is contrary to our purpose. This is because the supplementary labels from the baseline method are concentrated on the head class and are thus powerless in alleviating the long-tail problem. In contrast, Trico considers the correlations of labels more comprehensively and alleviates the long-tail problem of the dataset without hurting the prediction ability of the head labels. This is verified by the results, which show improvements of 2.08\% and 4.53\% in the body and tail group, respectively.

Based on our observations, we noticed that a majority of the labels supplemented using \textbf{S} (correlation of $\bm{A}_S$) belonged to the Body group, such as the labels \texttt{walk\_past} and \texttt{walk\_away} shown in Figure~\ref{fig:Results} (a). The supplementation of these labels led to a more significant improvement in the performance of the Body group. Similarly, we observed that using \textbf{T} to supplement missing labels for both the Body and Tail groups, and using \textbf{E} to supplement missing labels for the Body group, had similar effects.

We combined the three kinds of correlations to achieve the full Trico. According to the results, we can see that the proposed correlations complement and promote each other, further proving the effectiveness and reliability of using these three correlations.

\subsection{Human Evaluation}
We conducted a manual verification to assess the validity of the labels we added using Trico based on the three types of correlations. The results in Tables \ref{tabel:vidvrd-predcls}, \ref{tabel:vidor-predcls}, \ref{tabel:vidvrd-sgdet}, and \ref{tabel:vidor-sgdet} already showed that the added labels improved the performance of the models. However, to further prove their validity, we visualized missing labels along with their corresponding video segments and asked volunteers to vote on the correctness of the added labels.

\begin{figure}[t!]
    \centering
    \includegraphics[width=0.9\linewidth]{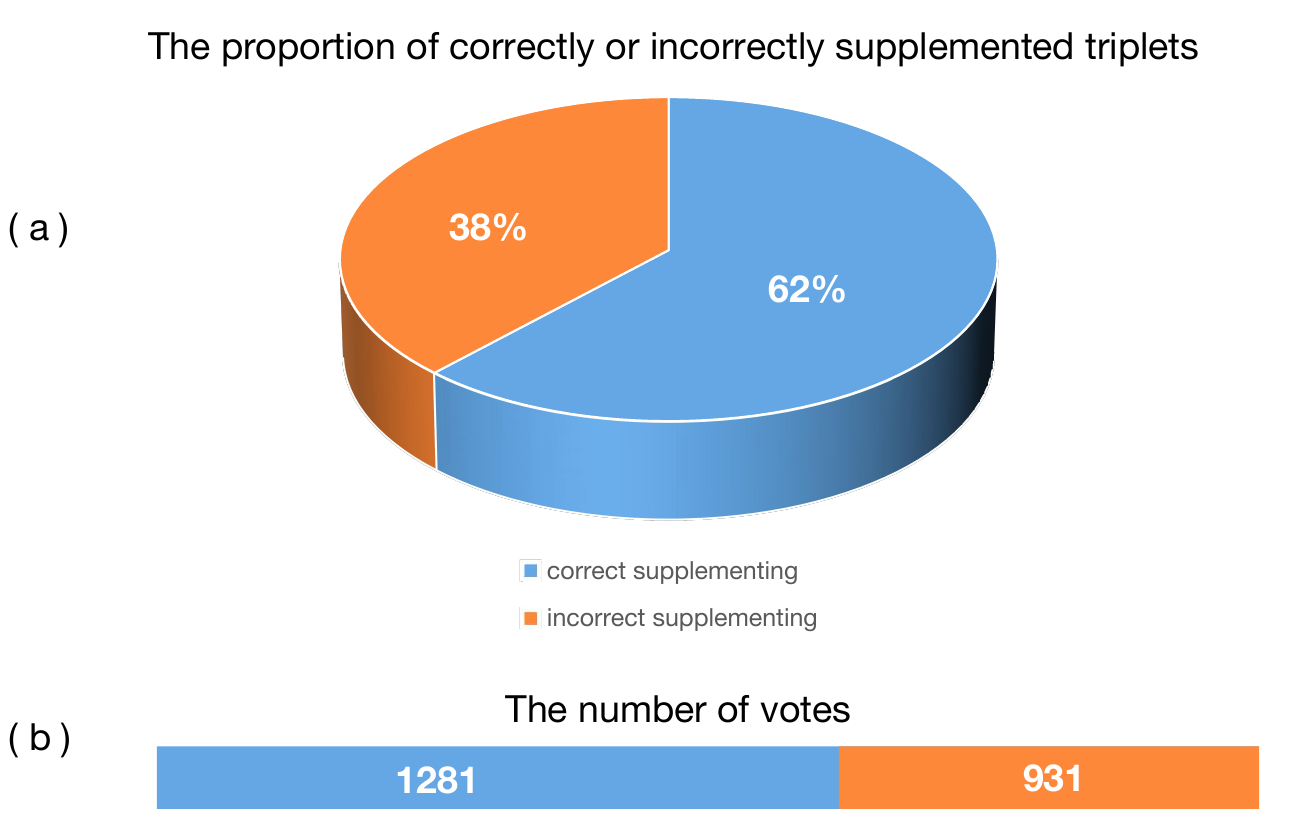} 
    \caption{\textbf{Statistical results of Human Evaluation. \textbf{(a)}: The proportion of correctly supplemented triplets. \textbf{(b)}: The number of votes counted, including correct and incorrect supplementing.}}
    \label{fig:statics}
\end{figure}

A total of $28$ volunteers were recruited, and $79$ additional triplet labels were randomly selected, resulting in a total of $2,212$ votes. Volunteers were given enough time to familiarize themselves with the video clips and understand the relevant content before voting. As shown in the figure~\ref{fig:statics}, results showed that 62\% of volunteers selected the correct number of people in the triplet label, indicating that our added labels were reasonable. Overall, we received $1,281$ votes for correct supplementing and $931$ votes for incorrect supplementing, further validating the effectiveness of our method.

\subsection{Qualitative Results}

\noindent{\textbf{Statistical results of the label supplementing on VidVRD.}} Figure~\ref{fig:Distribution} shows the statistics of the label supplementing results on the VidVRD~\cite{shang2017video} dataset. It can be observed that our supplemented labels significantly increase the proportion of tail labels in the entire dataset. This is important because tail labels are often underrepresented in datasets and can be difficult to learn. By supplementing these labels with Trico, we can improve the performance of models that rely on these datasets and make them more robust to real-world scenarios.

\noindent{\textbf{Visualization for the generated video scene graph on VidVRD.}} Figure~\ref{fig:Results} shows some video scene graphs generated after supplementing labels using Trico. Comparing them to the vanilla scene graph, our generated scene graphs are able to portray richer spatio-temporal relations of a dynamic scene. This is because Trico supplements the original labels in a way that captures more complex correlations between objects and relationships in a video. By visualizing these relationships in the form of a scene graph, we can better understand the dynamics of a video and extract more meaningful information from it.

\section{Conclusion}
In this paper, we focus on the VidSGG task, specifically addressing the biased distribution and missing annotation issues that inherently exist in the training data and hinder VidSGG performance. To address this problem, we propose \textbf{Trico}, the first method to approach VidSGG from an \emph{explicit} perspective of missing label supplementation. We explore triple complementary correlations to guide label supplementation. By capitalizing on the spatio-temporal cues offered by these correlations, missing labels can be effectively supplemented to achieve unbiased graph generation. The extensive results demonstrate the effectiveness of Trico and its state-of-the-art performance, particularly on tail predicates.

\section{Limitations} 
\begin{figure}[t!]
    \centering
    \includegraphics[width=\linewidth]{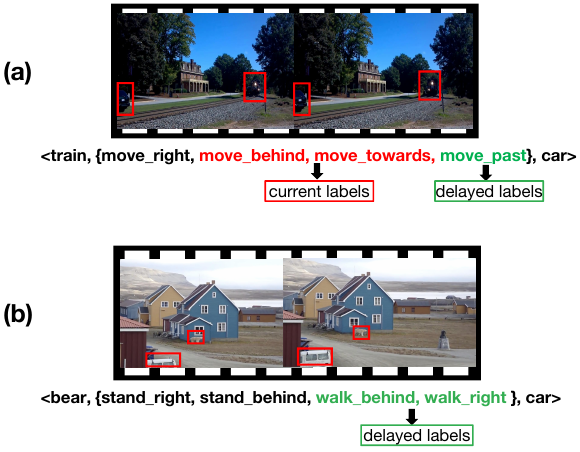} 
    \caption{\textbf{Supplementation examples of delayed labels. \textcolor{red}{Red}: current missing labels. \textcolor[rgb]{0.3,0.68,0.36}{Green}: delayed missing labels.}}
    \label{fig:discussion}
\end{figure}

In practice, we have observed that some of the predicate labels supplemented by temporal correlation correspond to events that are about to happen but have not yet occurred, as depicted in Figure~\ref{fig:discussion}. Although this is not entirely accurate, it is still consistent with the trend of the video and can provide useful cues for the model to generate video scene graphs. In future work, it would be interesting to investigate the impact of delayed labels and how to better handle such cases.

\bibliographystyle{ACM-Reference-Format}
\balance
\bibliography{mm}


\begin{thebibliography}{41}


\ifx \showCODEN    \undefined \def \showCODEN     #1{\unskip}     \fi
\ifx \showDOI      \undefined \def \showDOI       #1{#1}\fi
\ifx \showISBNx    \undefined \def \showISBNx     #1{\unskip}     \fi
\ifx \showISBNxiii \undefined \def \showISBNxiii  #1{\unskip}     \fi
\ifx \showISSN     \undefined \def \showISSN      #1{\unskip}     \fi
\ifx \showLCCN     \undefined \def \showLCCN      #1{\unskip}     \fi
\ifx \shownote     \undefined \def \shownote      #1{#1}          \fi
\ifx \showarticletitle \undefined \def \showarticletitle #1{#1}   \fi
\ifx \showURL      \undefined \def \showURL       {\relax}        \fi
\providecommand\bibfield[2]{#2}
\providecommand\bibinfo[2]{#2}
\providecommand\natexlab[1]{#1}
\providecommand\showeprint[2][]{arXiv:#2}

\bibitem[Antol et~al\mbox{.}(2015)]%
        {antol2015vqa}
\bibfield{author}{\bibinfo{person}{Stanislaw Antol}, \bibinfo{person}{Aishwarya
  Agrawal}, \bibinfo{person}{Jiasen Lu}, \bibinfo{person}{Margaret Mitchell},
  \bibinfo{person}{Dhruv Batra}, \bibinfo{person}{C~Lawrence Zitnick}, {and}
  \bibinfo{person}{Devi Parikh}.} \bibinfo{year}{2015}\natexlab{}.
\newblock \showarticletitle{Vqa: Visual question answering}. In
  \bibinfo{booktitle}{\emph{ICCV}}. \bibinfo{pages}{2425--2433}.
\newblock


\bibitem[Cao et~al\mbox{.}(2021)]%
        {20213}
\bibfield{author}{\bibinfo{person}{Q. Cao}, \bibinfo{person}{H. Huang},
  \bibinfo{person}{X. Shang}, \bibinfo{person}{B. Wang}, {and}
  \bibinfo{person}{T.~S. Chua}.} \bibinfo{year}{2021}\natexlab{}.
\newblock \showarticletitle{3-D Relation Network for visual relation
  recognition in videos}.
\newblock \bibinfo{journal}{\emph{Neurocomputing}}  \bibinfo{volume}{432}
  (\bibinfo{year}{2021}), \bibinfo{pages}{91--100}.
\newblock


\bibitem[Chen et~al\mbox{.}(2021)]%
        {Chen2021Social}
\bibfield{author}{\bibinfo{person}{Shuo Chen}, \bibinfo{person}{Zenglin Shi},
  \bibinfo{person}{Pascal Mettes}, {and} \bibinfo{person}{Cees G.~M. Snoek}.}
  \bibinfo{year}{2021}\natexlab{}.
\newblock \showarticletitle{Social Fabric: Tubelet Compositions for Video
  Relation Detection}. In \bibinfo{booktitle}{\emph{ICCV}}.
\newblock


\bibitem[Chen et~al\mbox{.}(2023)]%
        {chen2023video}
\bibfield{author}{\bibinfo{person}{Siqi Chen}, \bibinfo{person}{Jun Xiao},
  {and} \bibinfo{person}{Long Chen}.} \bibinfo{year}{2023}\natexlab{}.
\newblock \showarticletitle{Video scene graph generation from single-frame weak
  supervision}. In \bibinfo{booktitle}{\emph{The Eleventh International
  Conference on Learning Representations}}.
\newblock


\bibitem[Chen et~al\mbox{.}(2019)]%
        {chen2019knowledge}
\bibfield{author}{\bibinfo{person}{Tianshui Chen}, \bibinfo{person}{Weihao Yu},
  \bibinfo{person}{Riquan Chen}, {and} \bibinfo{person}{Liang Lin}.}
  \bibinfo{year}{2019}\natexlab{}.
\newblock \showarticletitle{Knowledge-embedded routing network for scene graph
  generation}. In \bibinfo{booktitle}{\emph{CVPR}}.
  \bibinfo{pages}{6163--6171}.
\newblock


\bibitem[Cong et~al\mbox{.}(2021)]%
        {cong2021spatial}
\bibfield{author}{\bibinfo{person}{Yuren Cong}, \bibinfo{person}{Wentong Liao},
  \bibinfo{person}{Hanno Ackermann}, \bibinfo{person}{Bodo Rosenhahn}, {and}
  \bibinfo{person}{Michael~Ying Yang}.} \bibinfo{year}{2021}\natexlab{}.
\newblock \showarticletitle{Spatial-temporal transformer for dynamic scene
  graph generation}. In \bibinfo{booktitle}{\emph{ICCV}}.
  \bibinfo{pages}{16372--16382}.
\newblock


\bibitem[Dong et~al\mbox{.}(2021)]%
        {dong2021dual}
\bibfield{author}{\bibinfo{person}{Jianfeng Dong}, \bibinfo{person}{Xirong Li},
  \bibinfo{person}{Chaoxi Xu}, \bibinfo{person}{Xun Yang},
  \bibinfo{person}{Gang Yang}, \bibinfo{person}{Xun Wang}, {and}
  \bibinfo{person}{Meng Wang}.} \bibinfo{year}{2021}\natexlab{}.
\newblock \showarticletitle{Dual encoding for video retrieval by text}.
\newblock \bibinfo{journal}{\emph{TPAMI}} (\bibinfo{year}{2021}).
\newblock


\bibitem[Feng et~al\mbox{.}(2021)]%
        {feng2021exploiting}
\bibfield{author}{\bibinfo{person}{Shengyu Feng}, \bibinfo{person}{Subarna
  Tripathi}, \bibinfo{person}{Hesham Mostafa}, \bibinfo{person}{Marcel Nassar},
  {and} \bibinfo{person}{Somdeb Majumdar}.} \bibinfo{year}{2021}\natexlab{}.
\newblock \showarticletitle{Exploiting Long-Term Dependencies for Generating
  Dynamic Scene Graphs}.
\newblock \bibinfo{journal}{\emph{arXiv preprint arXiv:2112.09828}}
  (\bibinfo{year}{2021}).
\newblock


\bibitem[Han et~al\mbox{.}(2016)]%
        {han2016seq}
\bibfield{author}{\bibinfo{person}{Wei Han}, \bibinfo{person}{Pooya Khorrami},
  \bibinfo{person}{Tom~Le Paine}, \bibinfo{person}{Prajit Ramachandran},
  \bibinfo{person}{Mohammad Babaeizadeh}, \bibinfo{person}{Honghui Shi},
  \bibinfo{person}{Jianan Li}, \bibinfo{person}{Shuicheng Yan}, {and}
  \bibinfo{person}{Thomas~S Huang}.} \bibinfo{year}{2016}\natexlab{}.
\newblock \showarticletitle{Seq-nms for video object detection}.
\newblock \bibinfo{journal}{\emph{arXiv preprint arXiv:1602.08465}}
  (\bibinfo{year}{2016}).
\newblock


\bibitem[Ji et~al\mbox{.}(2020)]%
        {ji2020action}
\bibfield{author}{\bibinfo{person}{Jingwei Ji}, \bibinfo{person}{Ranjay
  Krishna}, \bibinfo{person}{Li Fei-Fei}, {and} \bibinfo{person}{Juan~Carlos
  Niebles}.} \bibinfo{year}{2020}\natexlab{}.
\newblock \showarticletitle{Action genome: Actions as compositions of
  spatio-temporal scene graphs}. In \bibinfo{booktitle}{\emph{CVPR}}.
  \bibinfo{pages}{10236--10247}.
\newblock


\bibitem[Kingma and Ba(2014)]%
        {kingma2014adam}
\bibfield{author}{\bibinfo{person}{Diederik~P Kingma} {and}
  \bibinfo{person}{Jimmy Ba}.} \bibinfo{year}{2014}\natexlab{}.
\newblock \showarticletitle{Adam: A method for stochastic optimization}.
\newblock \bibinfo{journal}{\emph{arXiv}} (\bibinfo{year}{2014}).
\newblock


\bibitem[Li et~al\mbox{.}(2022)]%
        {li2022label}
\bibfield{author}{\bibinfo{person}{Lin Li}, \bibinfo{person}{Long Chen},
  \bibinfo{person}{Hanrong Shi}, \bibinfo{person}{Wenxiao Wang},
  \bibinfo{person}{Jian Shao}, \bibinfo{person}{Yi Yang}, {and}
  \bibinfo{person}{Jun Xiao}.} \bibinfo{year}{2022}\natexlab{}.
\newblock \showarticletitle{Label Semantic Knowledge Distillation for Unbiased
  Scene Graph Generation}.
\newblock \bibinfo{journal}{\emph{arXiv preprint arXiv:2208.03763}}
  (\bibinfo{year}{2022}).
\newblock


\bibitem[Li et~al\mbox{.}(2021b)]%
        {li2021bipartite}
\bibfield{author}{\bibinfo{person}{Rongjie Li}, \bibinfo{person}{Songyang
  Zhang}, \bibinfo{person}{Bo Wan}, {and} \bibinfo{person}{Xuming He}.}
  \bibinfo{year}{2021}\natexlab{b}.
\newblock \showarticletitle{Bipartite Graph Network with Adaptive Message
  Passing for Unbiased Scene Graph Generation}. In
  \bibinfo{booktitle}{\emph{CVPR}}.
\newblock


\bibitem[Li et~al\mbox{.}(2021a)]%
        {li2021interventional}
\bibfield{author}{\bibinfo{person}{Yicong Li}, \bibinfo{person}{Xun Yang},
  \bibinfo{person}{Xindi Shang}, {and} \bibinfo{person}{Tat-Seng Chua}.}
  \bibinfo{year}{2021}\natexlab{a}.
\newblock \showarticletitle{Interventional video relation detection}. In
  \bibinfo{booktitle}{\emph{ACM MM}}. \bibinfo{pages}{4091--4099}.
\newblock


\bibitem[Liao et~al\mbox{.}(2020)]%
        {liao2020ppdm}
\bibfield{author}{\bibinfo{person}{Yue Liao}, \bibinfo{person}{Si Liu},
  \bibinfo{person}{Fei Wang}, \bibinfo{person}{Yanjie Chen},
  \bibinfo{person}{Chen Qian}, {and} \bibinfo{person}{Jiashi Feng}.}
  \bibinfo{year}{2020}\natexlab{}.
\newblock \showarticletitle{Ppdm: Parallel point detection and matching for
  real-time human-object interaction detection}. In
  \bibinfo{booktitle}{\emph{CVPR}}. \bibinfo{pages}{482--490}.
\newblock


\bibitem[Liu et~al\mbox{.}(2020)]%
        {liu2020beyond}
\bibfield{author}{\bibinfo{person}{Chenchen Liu}, \bibinfo{person}{Yang Jin},
  \bibinfo{person}{Kehan Xu}, \bibinfo{person}{Guoqiang Gong}, {and}
  \bibinfo{person}{Yadong Mu}.} \bibinfo{year}{2020}\natexlab{}.
\newblock \showarticletitle{Beyond short-term snippet: Video relation detection
  with spatio-temporal global context}. In \bibinfo{booktitle}{\emph{CVPR}}.
  \bibinfo{pages}{10840--10849}.
\newblock


\bibitem[Liu et~al\mbox{.}(2021)]%
        {liu2021fully}
\bibfield{author}{\bibinfo{person}{Hengyue Liu}, \bibinfo{person}{Ning Yan},
  \bibinfo{person}{Masood~S Mortazavi}, {and} \bibinfo{person}{Bir Bhanu}.}
  \bibinfo{year}{2021}\natexlab{}.
\newblock \showarticletitle{Fully Convolutional Scene Graph Generation}. In
  \bibinfo{booktitle}{\emph{CVPR}}.
\newblock


\bibitem[Misra et~al\mbox{.}(2016)]%
        {misra2016seeing}
\bibfield{author}{\bibinfo{person}{Ishan Misra}, \bibinfo{person}{C
  Lawrence~Zitnick}, \bibinfo{person}{Margaret Mitchell}, {and}
  \bibinfo{person}{Ross Girshick}.} \bibinfo{year}{2016}\natexlab{}.
\newblock \showarticletitle{Seeing through the human reporting bias: Visual
  classifiers from noisy human-centric labels}. In
  \bibinfo{booktitle}{\emph{CVPR}}. \bibinfo{pages}{2930--2939}.
\newblock


\bibitem[Newell and Deng(2017)]%
        {newell2017pixels}
\bibfield{author}{\bibinfo{person}{Alejandro Newell} {and} \bibinfo{person}{Jia
  Deng}.} \bibinfo{year}{2017}\natexlab{}.
\newblock \showarticletitle{Pixels to Graphs by Associative Embedding}. In
  \bibinfo{booktitle}{\emph{NIPS}}.
\newblock


\bibitem[Qian et~al\mbox{.}(2019)]%
        {qian2019video}
\bibfield{author}{\bibinfo{person}{Xufeng Qian}, \bibinfo{person}{Yueting
  Zhuang}, \bibinfo{person}{Yimeng Li}, \bibinfo{person}{Shaoning Xiao},
  \bibinfo{person}{Shiliang Pu}, {and} \bibinfo{person}{Jun Xiao}.}
  \bibinfo{year}{2019}\natexlab{}.
\newblock \showarticletitle{Video relation detection with spatio-temporal
  graph}. In \bibinfo{booktitle}{\emph{ACM MM}}. \bibinfo{pages}{84--93}.
\newblock


\bibitem[Ren et~al\mbox{.}(2015)]%
        {ren2015faster}
\bibfield{author}{\bibinfo{person}{Shaoqing Ren}, \bibinfo{person}{Kaiming He},
  \bibinfo{person}{Ross~B Girshick}, {and} \bibinfo{person}{Jian Sun}.}
  \bibinfo{year}{2015}\natexlab{}.
\newblock \showarticletitle{Faster R-CNN: Towards Real-Time Object Detection
  with Region Proposal Networks}. In \bibinfo{booktitle}{\emph{NeurIPS}}.
\newblock


\bibitem[Shang et~al\mbox{.}(2019)]%
        {shang2019annotating}
\bibfield{author}{\bibinfo{person}{Xindi Shang}, \bibinfo{person}{Donglin Di},
  \bibinfo{person}{Junbin Xiao}, \bibinfo{person}{Yu Cao}, \bibinfo{person}{Xun
  Yang}, {and} \bibinfo{person}{Tat-Seng Chua}.}
  \bibinfo{year}{2019}\natexlab{}.
\newblock \showarticletitle{Annotating objects and relations in user-generated
  videos}. In \bibinfo{booktitle}{\emph{ICMR}}. \bibinfo{pages}{279--287}.
\newblock


\bibitem[Shang et~al\mbox{.}(2021)]%
        {shang2021video}
\bibfield{author}{\bibinfo{person}{Xindi Shang}, \bibinfo{person}{Yicong Li},
  \bibinfo{person}{Junbin Xiao}, \bibinfo{person}{Wei Ji}, {and}
  \bibinfo{person}{Tat-Seng Chua}.} \bibinfo{year}{2021}\natexlab{}.
\newblock \showarticletitle{Video Visual Relation Detection via Iterative
  Inference}. In \bibinfo{booktitle}{\emph{ACM MM}}.
  \bibinfo{pages}{3654--3663}.
\newblock


\bibitem[Shang et~al\mbox{.}(2017)]%
        {shang2017video}
\bibfield{author}{\bibinfo{person}{Xindi Shang}, \bibinfo{person}{Tongwei Ren},
  \bibinfo{person}{Jingfan Guo}, \bibinfo{person}{Hanwang Zhang}, {and}
  \bibinfo{person}{Tat-Seng Chua}.} \bibinfo{year}{2017}\natexlab{}.
\newblock \showarticletitle{Video visual relation detection}. In
  \bibinfo{booktitle}{\emph{ACM MM}}. \bibinfo{pages}{1300--1308}.
\newblock


\bibitem[Snoek et~al\mbox{.}(2009)]%
        {snoek2009concept}
\bibfield{author}{\bibinfo{person}{Cees~GM Snoek}, \bibinfo{person}{Marcel
  Worring}, {et~al\mbox{.}}} \bibinfo{year}{2009}\natexlab{}.
\newblock \showarticletitle{Concept-based video retrieval}.
\newblock \bibinfo{journal}{\emph{FOUND TRENDS INF RET}} \bibinfo{volume}{2},
  \bibinfo{number}{4} (\bibinfo{year}{2009}), \bibinfo{pages}{215--322}.
\newblock


\bibitem[Su et~al\mbox{.}(2020)]%
        {su2020video}
\bibfield{author}{\bibinfo{person}{Zixuan Su}, \bibinfo{person}{Xindi Shang},
  \bibinfo{person}{Jingjing Chen}, \bibinfo{person}{Yu-Gang Jiang},
  \bibinfo{person}{Zhiyong Qiu}, {and} \bibinfo{person}{Tat-Seng Chua}.}
  \bibinfo{year}{2020}\natexlab{}.
\newblock \showarticletitle{Video Relation Detection via Multiple Hypothesis
  Association}. In \bibinfo{booktitle}{\emph{ACM MM}}.
  \bibinfo{pages}{3127--3135}.
\newblock


\bibitem[Sun et~al\mbox{.}(2019)]%
        {sun2019video}
\bibfield{author}{\bibinfo{person}{Xu Sun}, \bibinfo{person}{Tongwei Ren},
  \bibinfo{person}{Yuan Zi}, {and} \bibinfo{person}{Gangshan Wu}.}
  \bibinfo{year}{2019}\natexlab{}.
\newblock \showarticletitle{Video visual relation detection via multi-modal
  feature fusion}. In \bibinfo{booktitle}{\emph{ACM MM}}.
  \bibinfo{pages}{2657--2661}.
\newblock


\bibitem[Tang et~al\mbox{.}(2020)]%
        {tang2020unbiased}
\bibfield{author}{\bibinfo{person}{Kaihua Tang}, \bibinfo{person}{Yulei Niu},
  \bibinfo{person}{Jianqiang Huang}, \bibinfo{person}{Jiaxin Shi}, {and}
  \bibinfo{person}{Hanwang Zhang}.} \bibinfo{year}{2020}\natexlab{}.
\newblock \showarticletitle{Unbiased scene graph generation from biased
  training}. In \bibinfo{booktitle}{\emph{CVPR}}. \bibinfo{pages}{3716--3725}.
\newblock


\bibitem[Tang et~al\mbox{.}(2019)]%
        {tang2019learning}
\bibfield{author}{\bibinfo{person}{Kaihua Tang}, \bibinfo{person}{Hanwang
  Zhang}, \bibinfo{person}{Baoyuan Wu}, \bibinfo{person}{Wenhan Luo}, {and}
  \bibinfo{person}{Wei Liu}.} \bibinfo{year}{2019}\natexlab{}.
\newblock \showarticletitle{Learning to compose dynamic tree structures for
  visual contexts}. In \bibinfo{booktitle}{\emph{CVPR}}.
  \bibinfo{pages}{6619--6628}.
\newblock


\bibitem[Tapaswi et~al\mbox{.}(2016)]%
        {tapaswi2016movieqa}
\bibfield{author}{\bibinfo{person}{Makarand Tapaswi}, \bibinfo{person}{Yukun
  Zhu}, \bibinfo{person}{Rainer Stiefelhagen}, \bibinfo{person}{Antonio
  Torralba}, \bibinfo{person}{Raquel Urtasun}, {and} \bibinfo{person}{Sanja
  Fidler}.} \bibinfo{year}{2016}\natexlab{}.
\newblock \showarticletitle{Movieqa: Understanding stories in movies through
  question-answering}. In \bibinfo{booktitle}{\emph{CVPR}}.
  \bibinfo{pages}{4631--4640}.
\newblock


\bibitem[Teng et~al\mbox{.}(2021)]%
        {teng2021target}
\bibfield{author}{\bibinfo{person}{Yao Teng}, \bibinfo{person}{Limin Wang},
  \bibinfo{person}{Zhifeng Li}, {and} \bibinfo{person}{Gangshan Wu}.}
  \bibinfo{year}{2021}\natexlab{}.
\newblock \showarticletitle{Target Adaptive Context Aggregation for Video Scene
  Graph Generation}. In \bibinfo{booktitle}{\emph{ICCV}}.
  \bibinfo{pages}{13688--13697}.
\newblock


\bibitem[Tsai et~al\mbox{.}(2019)]%
        {tsai2019video}
\bibfield{author}{\bibinfo{person}{Yao-Hung~Hubert Tsai},
  \bibinfo{person}{Santosh Divvala}, \bibinfo{person}{Louis-Philippe Morency},
  \bibinfo{person}{Ruslan Salakhutdinov}, {and} \bibinfo{person}{Ali Farhadi}.}
  \bibinfo{year}{2019}\natexlab{}.
\newblock \showarticletitle{Video relationship reasoning using gated
  spatio-temporal energy graph}. In \bibinfo{booktitle}{\emph{CVPR}}.
  \bibinfo{pages}{10424--10433}.
\newblock


\bibitem[Wei et~al\mbox{.}(2019)]%
        {wei2019neural}
\bibfield{author}{\bibinfo{person}{Yinwei Wei}, \bibinfo{person}{Xiang Wang},
  \bibinfo{person}{Weili Guan}, \bibinfo{person}{Liqiang Nie},
  \bibinfo{person}{Zhouchen Lin}, {and} \bibinfo{person}{Baoquan Chen}.}
  \bibinfo{year}{2019}\natexlab{}.
\newblock \showarticletitle{Neural multimodal cooperative learning toward
  micro-video understanding}.
\newblock \bibinfo{journal}{\emph{TIP}}  \bibinfo{volume}{29}
  (\bibinfo{year}{2019}), \bibinfo{pages}{1--14}.
\newblock


\bibitem[Woo et~al\mbox{.}(2021)]%
        {woo2021and}
\bibfield{author}{\bibinfo{person}{Sangmin Woo}, \bibinfo{person}{Junhyug Noh},
  {and} \bibinfo{person}{Kangil Kim}.} \bibinfo{year}{2021}\natexlab{}.
\newblock \showarticletitle{What and When to Look?: Temporal Span Proposal
  Network for Video Visual Relation Detection}.
\newblock \bibinfo{journal}{\emph{arXiv preprint arXiv:2107.07154}}
  (\bibinfo{year}{2021}).
\newblock


\bibitem[Xiao et~al\mbox{.}(2021)]%
        {xiao2021next}
\bibfield{author}{\bibinfo{person}{Junbin Xiao}, \bibinfo{person}{Xindi Shang},
  \bibinfo{person}{Angela Yao}, {and} \bibinfo{person}{Tat-Seng Chua}.}
  \bibinfo{year}{2021}\natexlab{}.
\newblock \showarticletitle{Next-qa: Next phase of question-answering to
  explaining temporal actions}. In \bibinfo{booktitle}{\emph{CVPR}}.
  \bibinfo{pages}{9777--9786}.
\newblock


\bibitem[Xu et~al\mbox{.}(2017)]%
        {xu2017scene}
\bibfield{author}{\bibinfo{person}{Danfei Xu}, \bibinfo{person}{Yuke Zhu},
  \bibinfo{person}{Christopher~B Choy}, {and} \bibinfo{person}{Li Fei-Fei}.}
  \bibinfo{year}{2017}\natexlab{}.
\newblock \showarticletitle{Scene graph generation by iterative message
  passing}. In \bibinfo{booktitle}{\emph{CVPR}}. \bibinfo{pages}{5410--5419}.
\newblock


\bibitem[Xu et~al\mbox{.}(2015)]%
        {xu2015show}
\bibfield{author}{\bibinfo{person}{Kelvin Xu}, \bibinfo{person}{Jimmy Ba},
  \bibinfo{person}{Ryan Kiros}, \bibinfo{person}{Kyunghyun Cho},
  \bibinfo{person}{Aaron Courville}, \bibinfo{person}{Ruslan Salakhudinov},
  \bibinfo{person}{Rich Zemel}, {and} \bibinfo{person}{Yoshua Bengio}.}
  \bibinfo{year}{2015}\natexlab{}.
\newblock \showarticletitle{Show, attend and tell: Neural image caption
  generation with visual attention}. In \bibinfo{booktitle}{\emph{ICML}}. PMLR,
  \bibinfo{pages}{2048--2057}.
\newblock


\bibitem[Xu et~al\mbox{.}(2022)]%
        {xu2022meta}
\bibfield{author}{\bibinfo{person}{Li Xu}, \bibinfo{person}{Haoxuan Qu},
  \bibinfo{person}{Jason Kuen}, \bibinfo{person}{Jiuxiang Gu}, {and}
  \bibinfo{person}{Jun Liu}.} \bibinfo{year}{2022}\natexlab{}.
\newblock \showarticletitle{Meta spatio-temporal debiasing for video scene
  graph generation}. In \bibinfo{booktitle}{\emph{ECCV}}. Springer,
  \bibinfo{pages}{374--390}.
\newblock


\bibitem[Yan et~al\mbox{.}(2020)]%
        {yan2020pcpl}
\bibfield{author}{\bibinfo{person}{Shaotian Yan}, \bibinfo{person}{Chen Shen},
  \bibinfo{person}{Zhongming Jin}, \bibinfo{person}{Jianqiang Huang},
  \bibinfo{person}{Rongxin Jiang}, \bibinfo{person}{Yaowu Chen}, {and}
  \bibinfo{person}{Xian-Sheng Hua}.} \bibinfo{year}{2020}\natexlab{}.
\newblock \showarticletitle{Pcpl: Predicate-correlation perception learning for
  unbiased scene graph generation}. In \bibinfo{booktitle}{\emph{Proceedings of
  the 28th ACM International Conference on Multimedia}}.
  \bibinfo{pages}{265--273}.
\newblock


\bibitem[Zellers et~al\mbox{.}(2018)]%
        {zellers2018neural}
\bibfield{author}{\bibinfo{person}{Rowan Zellers}, \bibinfo{person}{Mark
  Yatskar}, \bibinfo{person}{Sam Thomson}, {and} \bibinfo{person}{Yejin Choi}.}
  \bibinfo{year}{2018}\natexlab{}.
\newblock \showarticletitle{Neural motifs: Scene graph parsing with global
  context}. In \bibinfo{booktitle}{\emph{CVPR}}. \bibinfo{pages}{5831--5840}.
\newblock


\bibitem[Zheng et~al\mbox{.}(2019)]%
        {zheng2019relation}
\bibfield{author}{\bibinfo{person}{Sipeng Zheng}, \bibinfo{person}{Xiangyu
  Chen}, \bibinfo{person}{Shizhe Chen}, {and} \bibinfo{person}{Qin Jin}.}
  \bibinfo{year}{2019}\natexlab{}.
\newblock \showarticletitle{Relation understanding in videos}. In
  \bibinfo{booktitle}{\emph{ACM MM}}. \bibinfo{pages}{2662--2666}.
\newblock


\end{thebibliography}
\end{document}